\documentclass[final]{siamltex}
\usepackage{algorithm,algpseudocode}
\usepackage{latexsym, graphicx, epsfig, amsmath, amsfonts,amssymb,bm}
\usepackage{caption, subcaption}
\usepackage{epstopdf}
\usepackage{booktabs}
\usepackage{array}
\usepackage{color}
\usepackage{url}
\usepackage{comment}
\usepackage{diagbox}

\allowdisplaybreaks

\def\be{\begin{equation}}
\def\ee{\end{equation}}

\def\x{\mathbf{x}}

\def\f{\mathbf{f}}

\def\N{\mathbf{N}}

\def\PPh{\boldsymbol{\Phi}}
\def\PPs{\boldsymbol{\Psi}}

\def\R{{\mathbb R}}

\def\I{\mathbf I}

\def\A{\mathbf{A}}

\def\w{\mathbf{w}}

\def\0{\mathbf{0}}

\title{Learning Fine Scale Dynamics from Coarse Observations via Inner Recurrence}

\author{Victor Churchill\footnotemark[1]\thanks{Department of Mathematics,
		The Ohio State University, Columbus, OH 43210, USA. Emails:
		{\tt churchill.77@osu.edu, xiu.16@osu.edu} Funding: This 
		work was partially supported by AFOSR FA9550-22-1-0011.}\and Dongbin Xiu\footnotemark[1]
				}

\begin{document}
\maketitle
\begin{abstract}
Recent work has focused on data-driven learning of the evolution of unknown systems via deep neural networks (DNNs), with the goal of conducting long term prediction of the dynamics of the unknown system. In many real-world applications, data from time-dependent systems are often collected on a time scale that is coarser than desired, due to various restrictions during the data acquisition process. Consequently, the observed dynamics can be severely under-sampled and do not reflect the true dynamics of the underlying system. This paper presents a computational technique to learn the fine-scale dynamics from such coarsely observed data. The method employs inner recurrence of a DNN to recover the fine-scale evolution operator of the underlying system. In addition to mathematical justification, several challenging numerical examples, including unknown systems of both ordinary and partial differential equations, are presented to demonstrate the effectiveness of the proposed method.
\end{abstract}

\section{Introduction} \label{sec:intro}

Data-driven modeling and learning of unknown dynamical systems, including both ordinary and partial differential equations, has been a prominent area of research over the past few years. One approach to this problem, known as governing equation discovery, constructs a mapping from the state variables of the unknown system to their time derivatives. The equation itself is modeled using sparse approximation from a large dictionary of potential right-hand-side functions. In certain circumstances, exact equation recovery is possible. See, for example, \cite{brunton2016discovering} and the related work in recovering both ODEs (\cite{brunton2016discovering,kang2019ident, schaeffer2017sparse,schaeffer2017extracting, tran2017exact}) and PDEs (\cite{rudy2017data, schaeffer2017learning}). More recently, deep neural networks (DNNs) have also been used to construct this mapping. See, for example, \cite{lu2021deepxde,qin2018data,raissi2018multistep,rudy2018deep} for ODE modeling and \cite{long2018pde,long2017pde,lu2021learning,raissi2017physics1,raissi2017physics2,raissi2018deep,sun2019neupde} for PDE modeling.

An alternative approach, known as flow map or evolution operator discovery, seeks to model the unknown system by constructing an approximation to
its flow map over a short time (\cite{qin2018data}). Although the method does not directly recover the unknown governing equations,
it produces an accurate predictive model, once an accurate flow map approximation is constructed, so that any new initial condition can be accurately marched forward in time for long-term prediction. One advantage of flow map based learning is that it does not require approximations of temporal derivatives, which are prone to larger numerical errors. This approach uses DNNs, in particular residual networks (ResNet \cite{he2016deep}), as nonlinear approximators to model the flow map from data. Introduced to model autonomous systems in \cite{qin2018data}, this general framework of flow map based DNN learning has been extended to model non-autonomous systems \cite{QinChenJakemanXiu_SISC},  parametric dynamical systems \cite{QinChenJakemanXiu_IJUQ}, partially observed dynamical systems \cite{FuChangXiu_JMLMC20}, chaotic systems such as Lorenz 63 and 96 \cite{churchill2022deep}, as well as PDEs in modal space \cite{WuXiu_modalPDE}, and nodal space \cite{chen2022deep}. Adjustments have also been proposed to improve robustness through ensemble averaged learning \cite{churchill2022robust}. Relevant components of this framework are reviewed below.

As the field of unknown system learning matures, and examples start to be drawn from real-world applications rather than synthetic benchmarking problems, the methodology needs to keep pace with circumstances and limitations of data collection. One such scenario is that data are frequently collected on a coarser time scale than desired, which makes them appear discontinuous or discrete and obfuscates the underlying smooth fine time scale dynamics of the system. For example, storage or equipment limitations can restrict collection of certain climate data to daily or hourly intervals, while in reality the underlying continuous dynamics would better understood by collecting over much shorter intervals, e.g. every minute or even more frequently. It is then natural to wonder if it is possible to construct an accurate model for the smooth unknown dynamics over a much shorter time scale using the coarsely observed data.
In this paper, we answer this question affirmatively. Hence, the chief contribution of this paper is a systematic and rigorous examination of data-driven flow map approximation of unknown systems which are coarsely observed in time using an approachable and mathematically grounded DNN framework, a topic that (to the best of our knowledge) has not been explored in the literature. In particular, we propose using the concept of \emph{inner recurrence} to require the network approximation of the flow map to operate on a finer time scale than the observations. This work builds upon the existing DNN framework for evolution discovery and opens up the possibility of addressing a more complex and broader class of problems. After presenting the DNN structure and training specifics, along with its mathematical justification, we present six challenging ODE and PDE learning problems to demonstrate the effectiveness of the method.
\section{Flow Map Modeling of Unknown Systems} \label{sec:setup}

We are interested in constructing accurate approximate models for the evolution
laws behind dynamical data.
%
Throughout this paper our discussion will be on unknown systems observed at discrete time
instances with a constant time step $\Delta$,
\be \label{tline}
t_0<t_1<\cdots, \qquad
t_{n+1} - t_n = \Delta, \quad \forall n.
\ee
Generality is not lost with the constant time step assumption. 
We will also use a subscript to denote the time variable of
a function, e.g., $\x_n = \x(t_n)$. Some parts of this review closely follow that of \cite{churchill2022robust}.



Consider an unknown autonomous system
\be\label{eq:ODE}
\frac{d\x}{dt} = \f(\x), \qquad \x\in\R^d,
\ee
where $\f:\R^d\to \R^d$ is unknown. Because the system is autonomous, its flow map depends only
on the time difference but not the actual time, i.e.,
$ \x_n = \PPh_{t_n-t_s}(\x_s)$. Thus, the solution over one time step
satisfies
\be
\x_{n+1} = \PPh_{\Delta}(\x_n) = \x_n + \PPs_{\Delta}(\x_n),
\ee
where $\PPs_{\Delta} = \PPh_{\Delta} - \mathbf{I}$, with $\mathbf{I}$ as the
identity operator.

\begin{figure}[t]
	\begin{center}
		\includegraphics[width=\textwidth]{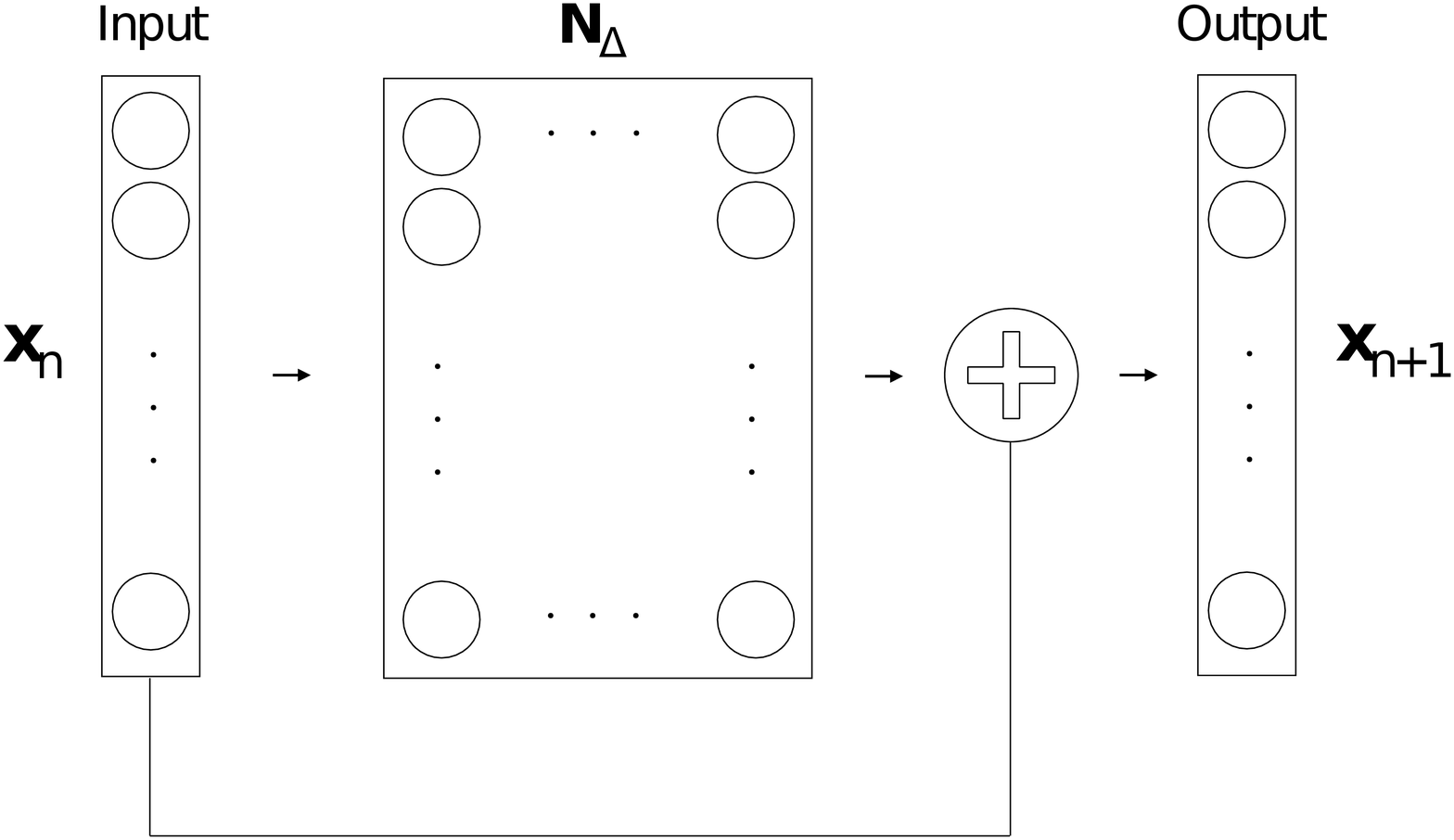}
		\caption{ResNet DNN Structure.}
		\label{fig:structure}
	\end{center}
\end{figure}

When data for the state variables $\x$ over the time stencil
\eqref{tline} are available, they can be grouped into pairs separated by
one time step
$$
\{\x^{(m)}(0), ~~\x^{(m)}(\Delta)\},\quad m=1,\ldots,M,
$$
where $M$ is the total number of such data pairs. This is the training data set.
One can define a residual network (ResNet \cite{he2016deep}) in the
form of
\be
\mathbf{y}^{out} = \left[\mathbf{I}+\mathbf{N}_\Delta \right](\mathbf{y}^{in}),
\ee
where $\N_\Delta:\R^d\to\R^d$ stands for the mapping operator of a standard
feedforward fully connected neural network. An illustration of this structure is shown in Figure \ref{fig:structure}.
The network is then trained by using the training data set 
and minimizing the mean squared loss function
\be
\frac1M\sum_{m=1}^M \left\| \x^{(m)}(\Delta) - [\I+\N_\Delta](\x^{(m)}(0);\Theta)\right\|^2,
\ee
with respect to the hyperparameters of the network (weights and
biases) $\Theta$. If the network is well-trained then
$\N_\Delta\approx \PPs_\Delta$.

For predictions, numerical stability is essential. Although
theoretical analysis is not available at the moment, extensive
numerical experiments have shown that numerical stability can be
significantly enhanced by using a loss function with \emph{outer
  recurrence}.
That is, if a sequence of $R_{out}$ snapshots
\be\label{observations}
\{\x^{(m)}(0),\x^{(m)}(\Delta),\ldots,\x^{(m)}(R_{out}\Delta)\},\quad m=1,\ldots,M,
\ee
are collected as training data, and the recurrent loss function
\be
\frac1M\sum_{m=1}^M \sum_{r=1}^{R_{out}}  \left\|\x^{(m)}(r\Delta) - [\mathbf{I}+\mathbf{N}_\Delta]^{r}(\x^{(m)}(0);\Theta)\right\|^2
\ee
is used instead, where $[\I+\N_\Delta]^r$ indicates composition of the network function with itself $r$ times.

The trained network thus accomplishes
$$
\x^{(m)}(r\Delta) \approx [\I+\N_\Delta]^r(\x^{(m)}(0)), \qquad \forall m=1,\ldots,M,~r=1,\ldots,R_{out}.
$$
Once the network is trained to satisfactory accuracy, it can be used as a predictive model on the $\Delta$ time scale
\be \label{ResNet}
\x_{n+1} =  \x_n + \N_\Delta(\x_n), \qquad n=0,1,\dots,
\ee
for any initial condition $\mathbf{x}(t_0)$. This framework was
proposed in  \cite{qin2018data}, with extensions to parametric systems
and time-dependent (non-autonomous) systems (\cite{QinChenJakemanXiu_IJUQ,QinChenJakemanXiu_SISC}).

\section{Learning Fine Scale Dynamics via Inner
  Recurrence} \label{sec:recurrence}

We now present the modeling of unknown systems on a finer time scale.
Assume, as in \eqref{observations}, that the training data set is again made up of observations of the state variables $\x$ of an unknown dynamical system at $t=0,\Delta,\ldots,R_{out}\Delta$.
Unlike in the previous section, however, the time scale $\Delta$ is not sufficient (i.e. too large) to describe and properly determine the dynamics of the underlying system. This could occur for many reasons, including e.g. limited storage resources, that make $\Delta$ insufficient to represent the smooth dynamics of the system. Hence we wish to learn the missing fine-scale time dynamics.

\subsection{Inner Recurrence for ODE modeling}

\begin{figure}[t]
	\begin{center}
		\includegraphics[width=\textwidth]{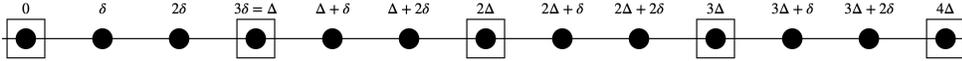}
		\caption{Time sampling diagram for $R_{in}=3$ and $R_{out}=4$. The squares represent the coarse observed time scale, while the circles represent the fine learning time scale.}
		\label{fig:sampling_diagram}
	\end{center}
\end{figure}

\begin{figure}[t]
	\begin{center}
		\includegraphics[width=\textwidth]{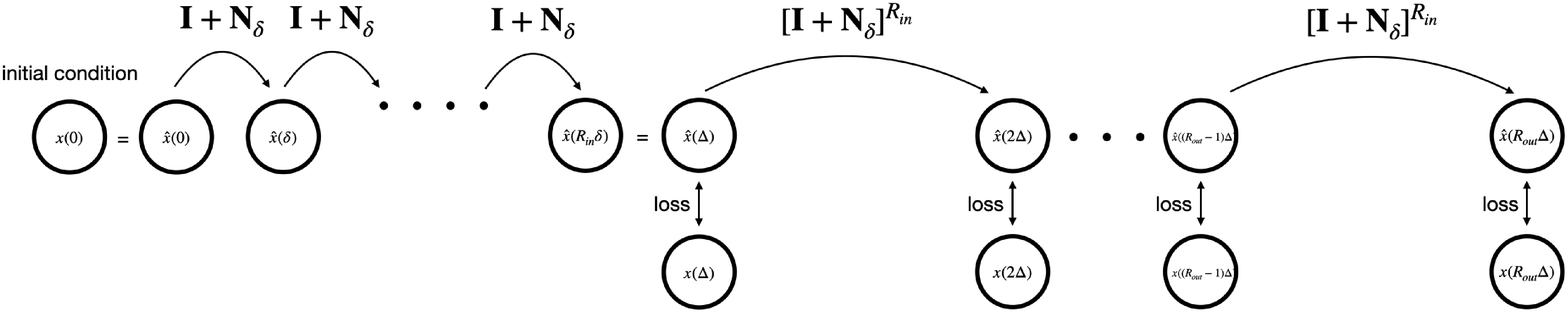}
		\caption{Loss diagram with inner and outer recurrence.}
		\label{fig:loss_diagram}
	\end{center}
\end{figure}

Therefore, we introduce the concept of \emph{inner recurrence} to learn the dynamics on a finer time scale $\delta<\Delta$. Inner recurrence dictates the time scale of the neural network function itself, that is, the length of time marched forward by the approximate flow map. No loss is measured at the inner recurrence steps, as there is no data for this fine time scale. However, we enforce that the network operator must compose $R_{in}$ times between observations such that $\N_\delta\approx \PPs_\delta$ (as opposed to $\PPs_\Delta$ as above). This is achieved by recurring the network function $R_{in}$ times (on the fine time scale $\delta$) in between when the loss is measured on the coarse time scale $\Delta$ for $r=1,\ldots,R_{out}$. In particular, the loss is now given by
\be \label{eq:fineloss}
\frac1M\sum_{m=1}^M \sum_{r=1}^{R_{out}}  \left\|\x^{(m)}(r\Delta) - [\I+\N_\delta]^{r\cdot R_{in}}(\x^{(m)}(0);\Theta)\right\|^2,
\ee
where $\delta$ is chosen such that $R_{in} = \Delta/\delta$ is an integer, and $[\I+\N_\delta]^{r\cdot R_{in}}$ represents the composition of the neural network function $r\cdot R_{in}$ times. In practice, the application and variation of the underlying function will dictate the appropriate size of $\delta$ and $R_{in}$ depending on the problem and the observation time scale $\Delta$. As illustrated in Figure \ref{fig:sampling_diagram}, if inner recurrence $R_{in}=3$ and outer recurrence $R_{out}=4$ then data is observed on the time scale $\Delta$ but the network learns on the time scale $\delta$. In training, the loss is minimized with respect to the network hyperparameters $\Theta$ via Adam, \cite{kingma2014adam}. Another diagram visualizes the loss computation in Figure \ref{fig:loss_diagram}.

Applying the learning of fine-scale time dynamics to the modeling of unknown systems is straightforward. In particular, the implementation only requires composing the network with itself $R_{in}$ times in between when the loss is measured, which effectively enforces the approximation of $\PPs_\delta$ as opposed to $\PPs_\Delta$. An important assumption in the approximation of a flow map $\PPs_\delta$ is that the time step $\Delta$ is small enough that there exists a unique flow map $\PPs_\Delta$. Later, we show an example where the observation time scale $\Delta$ is too large such that $\PPs_\Delta$ is not unique and therefore the network (depending on random initialization and stochastic optimization) finds either the appropriate map or another solution which goes through the same observation points. Therefore, we use the rule of thumb that if $\Delta$ is small enough (which is problem-dependent) then we can proceed as above.

\subsection{Inner Recurrence for PDE Modeling in Nodal Space}\label{sec:PDE}

\begin{figure}[t]
	\begin{center}
		\includegraphics[width=\textwidth]{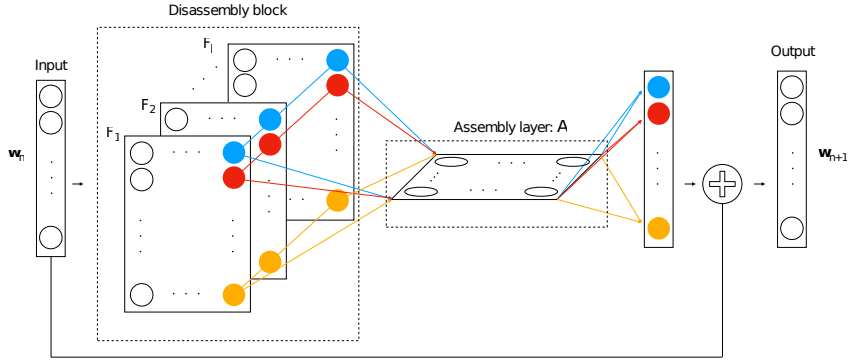}
		\caption{PDE DNN Structure.}
		\label{fig:pde_structure}
	\end{center}
\end{figure}

The flow map learning approach has also been applied in modeling PDEs. When the
solution of the PDE can be expressed using a fixed basis, the learning
can be conducted in modal space. In this case, the ResNet approach can
be adopted. See \cite{WuXiu_modalPDE} for details.
When data of the PDE solutions are available as nodal values over a
set of grids in physical space, its DNN learning is more involved. In this
case, a DNN structure was developed in \cite{chen2022deep}, which can
accommodate the situation when the data are on unstructured grids. It is based on a numerical scheme for solving the PDE, and 
consists of a set of specialized layers including disassembly layers and an
assembly layer, which are used to model the potential differential
operators involved in the unknown PDE.
The proposed DNN model defines the following mapping,
\be \label{NN_simple}
\w_{n+1} = \w_n + \A(\mathbf{F}_1(\w_n), \dots, \mathbf{F}_J(\w_n)),
\ee
where $\mathbf{F}_1,\dots, \mathbf{F}_J$ are the NN operators for the disassembly layers
and $\A$ is the NN operator for the assembly layer which operates componentwise on the grid elements. An illustration of this structure is shown in Figure \ref{fig:pde_structure}. The DNN modeling approach was shown to be highly flexible and accurate to learn a variety of PDEs in \cite{chen2022deep}. See \cite{chen2022deep} for more details on this structure and its mathematical properties. Since this configuration can also be viewed as a residual network application, i.e. $\w_{n+1} = [\I+\N](\w_n)$ where $\N = \A\circ(\mathbf{F}_1,\ldots,\mathbf{F}_J)$, we can directly apply the above inner and outer recurrence scheme described above.

\section{Computational Studies} \label{sec:examples}

In this section, we present six numerical examples (four for ODEs and two for PDEs) to demonstrate the properties of the proposed approach for learning fine scale time dynamics from coarse scale time observations using DNNs with inner recurrence.

\subsection{Computational Setting}
The structure of the DNNs used to achieve fine time scale flow map learning are those of \eqref{ResNet} and \eqref{NN_simple} depending on whether the underlying system is an ODE or PDE, which is reviewed in the previous sections, 
where Figures \ref{fig:structure} and \ref{fig:pde_structure} illustrate the two network structures. Our numerical experimentation and previous work with this framework indicate that particularly wide or deep networks are not typically necessary for learning with this structure. All specifications such as number of layers and neurons are included with each example.

For benchmarking purposes, in all examples the true systems we seek to approximate are in fact known. However, these true models serve only two purposes: (1) to generate synthetic data with which to train the DNN flow map approximations; and (2) to generate reference solutions for comparison with DNN predictions in testing. Therefore, the knowledge of the true system does not in any way facilitate the DNN model approximation. Data generation for both of these tasks is achieved by solving the true systems using a high-order numerical solver, and observing this reference solution at discrete time steps of length $\Delta$. To generate the training data, $M$ initial conditions generate trajectories from which a sequence of length $R_{out}+1$ is collected to form the training data set \eqref{observations}. In our examples, typically $M=10,000$ and $R_{out} \in \{1,\ldots,10\}$.

Once the training dataset has been generated, the learning task is achieved by training the DNN model with the data set \eqref{observations}. In particular, the network hyperparameters (weights and biases) are trained by minimizing the recurrent loss function \eqref{eq:fineloss} using the stochastic optimization method Adam \cite{kingma2014adam}. Typically the models are trained for for $10,000$ epochs with batch size $50$ and a constant learning rate of $10^{-3}$ in Tensorflow \cite{tensorflow2015}.

After satisfactory network training, we obtain a predictive model for the unknown system which can be marched forward in time on the $\delta$ time scale from any new initial condition. To validate the network prediction, testing data is generated in the same manner as training data was above. In particular, \emph{new} initial conditions (i.e. ones not fed through the network in training) generate reference solutions using the true governing equations. For prediction, the DNN model is marched forward starting with the first time step of the test trajectory and is compared against the reference. Note that we march forward significantly longer than $R_{out}\Delta$ (the length of each training sequence) to examine the long-term system behavior.

\subsection{Example 1: van der Pol Oscillator}

We first consider the 2nd order van der Pol oscillator,
\[ 
	\ddot{y}_1 - (1-y_1^2)\dot{y}_1+y_1 = 0,
\]
which can be rewritten as the two-dimensional system
\[ \begin{cases} 
	\dot{y}_1 = y_2,\\
	\dot{y}_2 = (1-y_1)^2y_2-y_1.
   \end{cases}
 \]
Training data are collected from the two variable system by sampling $10,000$ initial conditions uniformly over the computational domain of $[-2,2]\times[-1.5,1.5]$, solving the system numerically, and observing at time step $\Delta=2$ for a total time length of $T=20$. A standard ResNet with 3 hidden layers with 20 nodes each is used, with inner recurrence $R_{in} = 10$ and outer recurrences $R_{out} = \{1,10\}$ such that the fine time learning occurs on the scale of $\delta = \Delta/R_{in} = 0.2$. The mean squared loss function is minimized using Adam with a constant learning rate of $10^{-3}$ for $10,000$ epochs.

\begin{figure}[htbp]
	\begin{center}
		\includegraphics[width=\textwidth]{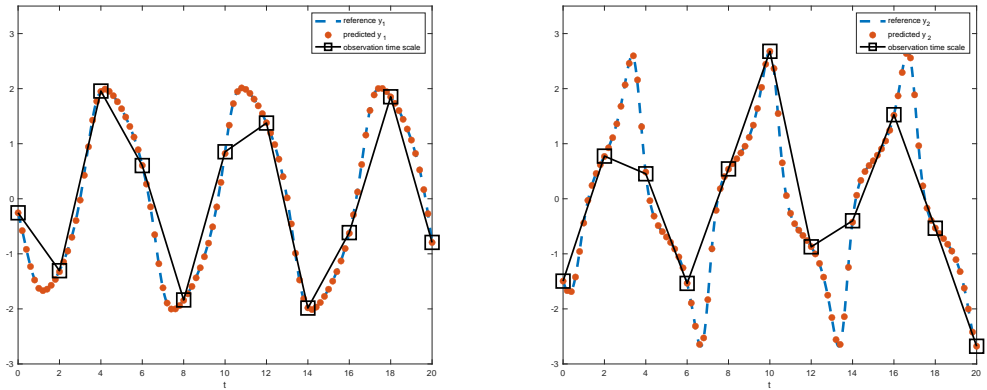}
		\caption{Ex. 1: van der Pol Oscillator -- Example test trajectory.}
		\label{fig:vanderpol_traj}
	\end{center}
\end{figure}

\begin{figure}[htbp]
	\begin{center}
		\includegraphics[width=\textwidth]{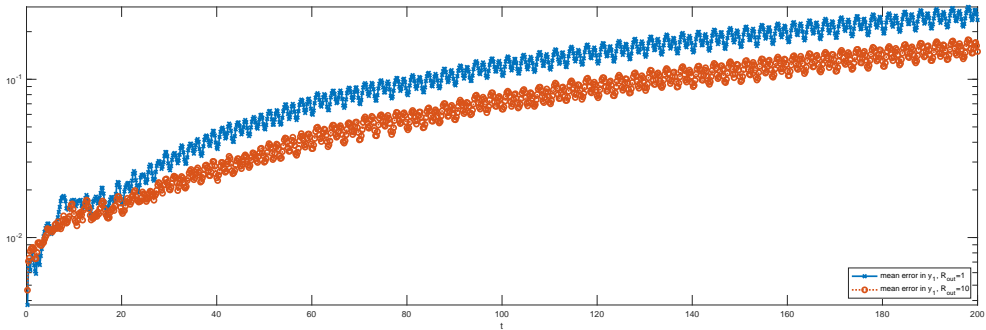}
		\caption{Ex. 1: van der Pol Oscillator -- Mean error over 100 test trajectories.}
		\label{fig:vanderpol_error}
	\end{center}
\end{figure}

Prediction is carried out for $T=200$ via $\delta=0.2$ ($1000$ time steps) using $100$ new initial conditions uniformly sampled from the same domain. In Figure \ref{fig:vanderpol_traj}, we see that despite the jagged appearance of the coarse time observations, the network accurately fills in the missing fine time scale dynamics. In Figure \ref{fig:vanderpol_error}, we look at the log of the average error of 100 test trajectories over time and notice that changing the outer recurrence $R_{out}$ from $1$ to $10$ has a positive effect on reducing long-term prediction error.

\subsection{Example 2: Violating conditions}

Next we consider the following pendulum problem,
\[ 
      \ddot{y}_1 = -\beta \sin y_1,
\]
which can be rewritten as the two-dimensional first order system
\[ \begin{cases} 
      \dot{y}_1 = y_2,\\
      \dot{y}_2 = -\beta \sin y_1,
   \end{cases}
\]
where $\beta=9.80665$. Training data are collected by sampling $10,000$ initial conditions uniformly over the computational domain of $[-\pi/2,\pi/2]\times[\pi,\pi]$, solving the system numerically, and observing at time step $\Delta=1$ for a total time length of $T=10$. A standard ResNet with 3 hidden layers with 20 nodes each is used, with outer recurrence $R_{out} = 1$ and inner recurrence $R_{in} = 10$ such that the fine time learning occurs on the scale of $\delta = \Delta/R_{in} = 0.1$. The mean squared loss function is minimized using Adam with a constant learning rate of $10^{-3}$ for $10,000$ epochs.

\begin{figure}[htbp]
	\begin{center}
		\includegraphics[width=\textwidth]{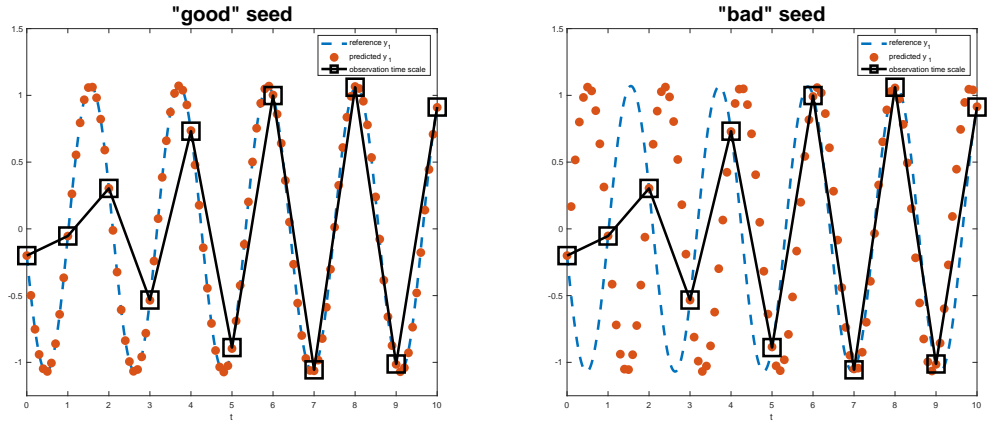}
		\caption{Ex. 2: Pendulum - Example test trajectory using two trained networks.}
		\label{fig:oscillator}
	\end{center}
\end{figure}

Two networks are trained using identical configurations but different random seeds. The random seed controls the initialization and stochastic optimization. In this case the size of $\Delta$ is too large, such that $\PPs_\Delta$ is not unique. That is, there are multiple paths between the observed data points. Hence, Figure \ref{fig:oscillator} shows that in prediction one seed finds the ``correct'' path through the new testing data while the other seed finds another path that is equally accurate at the observed data but does not represent the flow map governing the data generation. This shows that although the learning occurs on the time scale of $\delta$, it is still critical to have the observation time scale $\Delta$ small enough so that the problem is uniquely determined.

\subsection{Example 3: Differential-Algebraic System}

Next we consider a system of nonlinear differential-algebraic equations, a model for an electric network from \cite{pulch2013polynomial} also studied in \cite{chen2021generalized},
\[ \begin{cases} 
	\dot{u}_1 = v_2/C,\\
	\dot{u}_2 = u_1/L,\\
	0 = v_1 - (G_0-G_\infty)U_0\tanh(u_1/U_0)-G_\infty u_1,\\
	0 = v_2+u_2+v_1.
   \end{cases}
\]
where $u_1$ denotes the node voltage, while $u_2$ and $v_1$ are branch currents. The physical parameters are specified as $C=10^{-9}$, $L=10^{-6}$, $U_0=1$, $G_0=-0.1$, and $G_\infty = 0.25$. Training data are collected from the two variable system for $(u_1,u_2)$ by sampling $10,000$ initial conditions uniformly over the computational domain of $[-2,2]\times[-0.2,0.2]$, solving the system numerically, and observing at time step $\Delta=5\times10^{-8}$ for a total time length of $T=10^{-6}$. A standard ResNet with 3 hidden layers with 20 nodes each is used, with outer recurrence $R_{out} = 1$ and inner recurrence $R_{in} = 10$ such that the fine time learning occurs on the scale of $\delta = \Delta/R_{in} = 5\times10^{-9}$. The mean squared loss function is minimized using Adam with a constant learning rate of $10^{-3}$ for $10,000$ epochs.

\begin{figure}[htbp]
	\begin{center}
		\includegraphics[width=\textwidth]{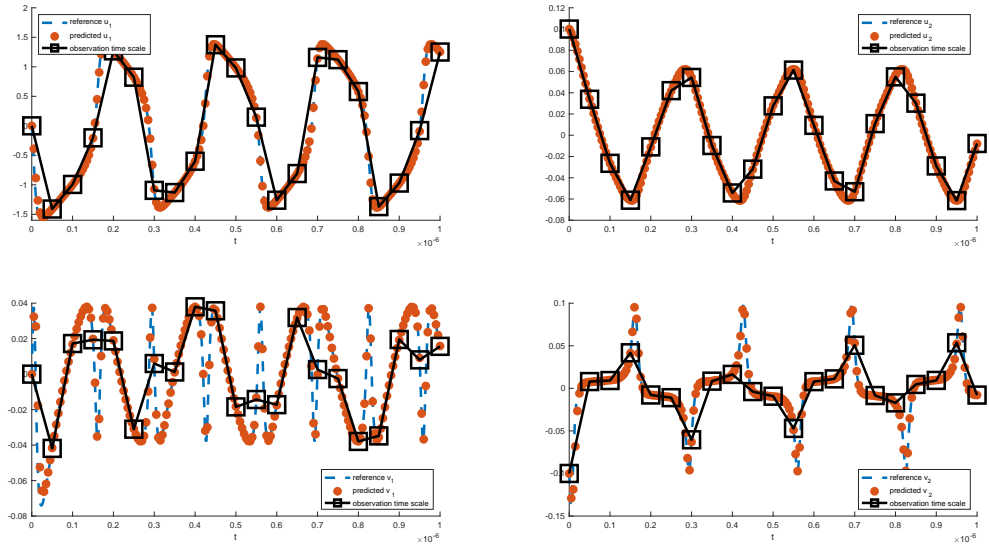}
		\caption{Ex. 3: Differential-Algebriac Equations - Example test trajectory.}
		\label{fig:dae}
	\end{center}
\end{figure}

Prediction is carried out to $T=10^{-6}$ via $\delta=5\times10^{-9}$ ($200$ time steps) using $100$ new initial conditions uniformly sampled from the same domain. The results are shown in Figure \ref{fig:dae}. We see from this example test trajectory that the network is still able to perform very accurately on the algebraic variables despite the coarse observations at times missing entire spikes of the functions.

\subsection{Example 4: Lorenz System}

We consider the Lorenz 63 system,
\[ \begin{cases} 
	\dot{x} = 10(y-x),\\
	\dot{y} = x(28-z)-y,\\
	\dot{z} = xy-\frac83 z,
   \end{cases}
\]
a three-dimensional nonlinear, deterministic, chaotic system. As in \cite{churchill2022deep}, which focuses on ResNet learning of the physics of chaotic systems, training data are collected from the full system by sampling $10,000$ chunks uniformly from a single solution of the system starting from initial condition $(1,1,1)$ with time step $\Delta=0.1$ for a total time length of $T=10,000$. A standard ResNet with 3 hidden layers with 20 nodes each is used, with outer recurrence $R_{out} = 1$ and inner recurrence $R_{in} = 10$ such that the fine time learning occurs on the scale of $\delta = \Delta/R_{in} = 0.01$. The mean squared loss function is minimized using Adam with a constant learning rate of $10^{-3}$ for $10,000$ epochs.

\begin{figure}[htbp]
	\begin{center}
		\includegraphics[width=\textwidth]{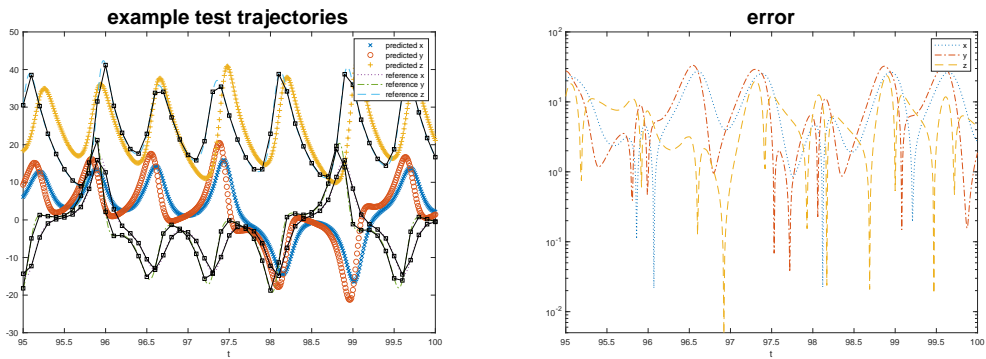}
		\caption{Ex. 4: Lorenz System -- Example test trajectory and error.}
		\label{fig:lorenz63coarse_error}
	\end{center}
\end{figure}

Prediction is carried out for $T=100$ via $\delta=0.01$ ($10,000$ time steps) using the new initial condition $(10,10,20)$ as an example. The results are shown in Figure \ref{fig:lorenz63coarse_error}. As in \cite{churchill2022deep}, we highlight that long term pointwise accuracy for chaotic systems is nearly impossible given the sensitivity of the true system behavior to perturbation and the fact that the system we learn is merely an approximation. Hence, we focus on other qualitative measurements such as bounded pointwise error, histogram and autocorrelation function matching, qualitatively similar phase plots, and measures that quantitatively measure chaos including correlation dimension \cite{theiler1987efficient}, approximate entropy \cite{pincus1991approximate}, and Lyapunov exponent \cite{rosenstein1993practical}, to demonstrate that the fine-scale chaotic behavior of the Lorenz system can be learned from coarser observations. These tools have been used to classify chaotic behavior in Lorenz and other chaotic systems, e.g. in \cite{bakker2000learning,chattopadhyay2020data,kim1999nonlinear,rudy2019deep}. All metrics are computed using MATLAB \cite{MATLAB}, in particular the Econometrics and Predictive Maintenance \cite{matlabpredmaint} toolboxes. See Section 3 of \cite{churchill2022deep} for computational details.

\begin{figure}[htbp]
	\begin{center}
		\includegraphics[width=\textwidth]{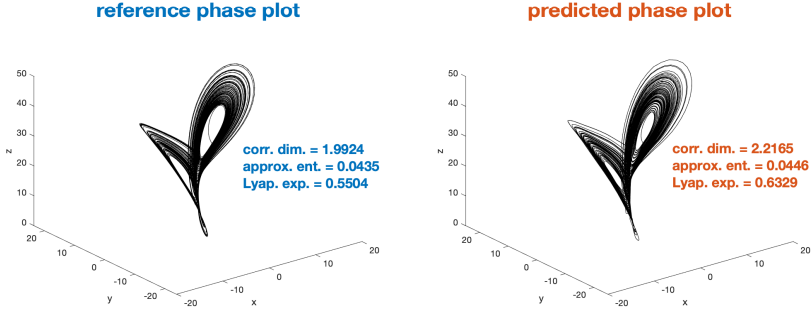}
		\caption{Ex. 4: Lorenz System -- Phase plots.}
		\label{fig:lorenz63coarse_phase}
	\end{center}
\end{figure}

\begin{figure}[htbp]
	\begin{center}
		\includegraphics[width=\textwidth]{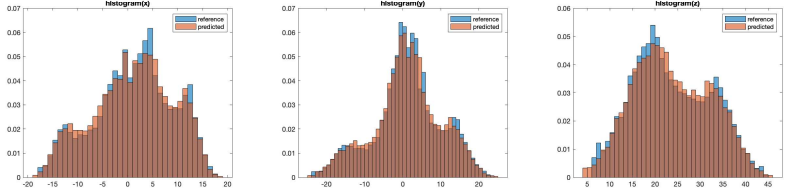}
		\caption{Ex. 4: Lorenz System -- Histograms.}
		\label{fig:lorenz63coarse_hist}
	\end{center}
\end{figure}

\begin{figure}[htbp]
	\begin{center}
		\includegraphics[width=\textwidth]{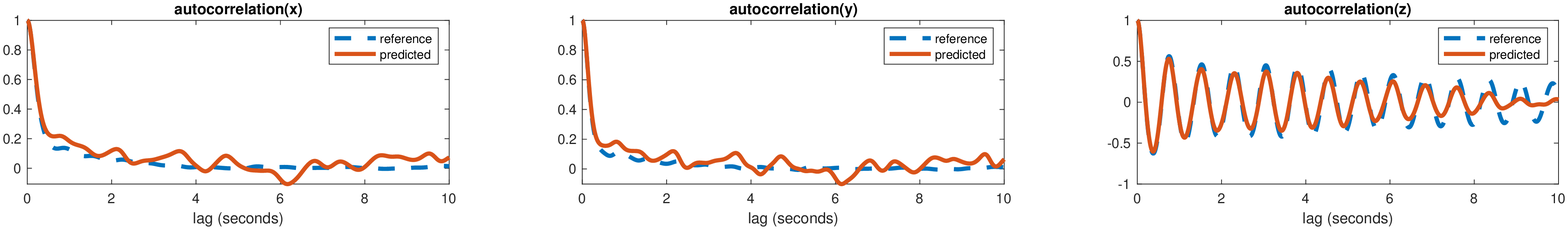}
		\caption{Ex. 4: Lorenz System -- Autocorrelation functions.}
		\label{fig:lorenz63coarse_autocorr}
	\end{center}
\end{figure}

Figure \ref{fig:lorenz63coarse_error} shows the pointwise error of the prediction between $t=95$ and $t=100$ seconds. We see that by this point in time the system prediction is far off from the reference, the chaotic behavior is replicated even on the fine time scale and the error even after this long time marching is bounded which demonstrates the long term stability of the prediction. Figure \ref{fig:lorenz63coarse_phase} shows that the phase plots for the reference is qualitatively similar to that of the prediction, and also quantitatively demonstrates the accuracy by displaying the chaos complexity measures mentioned above. Figure \ref{fig:lorenz63coarse_hist} shows a strong match to the approximate density of the values of the reference solution, while finally Figure \ref{fig:lorenz63coarse_autocorr} shows a good match to the reference autocorrelation functions in each variable.

\subsection{Example 5: One-dimensional PDE}

Moving to PDEs, we consider the one-dimensional FitzHugh-Nagumo systems of equations with diffusion,
\[ \begin{cases} 
	\frac{\partial v}{\partial t} = v-\frac13 v^3-w+D\nabla w,\\
	\frac{\partial w}{\partial t} = \epsilon(v+b-cw),
   \end{cases}
\]
with periodic boundary conditions in the computational domain $[0,5]$ discretized with $50$ points, 
where $D=0.01$, $\epsilon=0.08$, $b=0.7$, $c=0.8$. These reaction-diffusion equations simulate the propagation of waves in an excitable media, such as in heart tissue or nerve fiber. Training data are collected from the two variable system by sampling $10,000$ initial conditions
\begin{align*}
v(x,0) &\sim\begin{cases}       U[0.9,1.1] & m_1\leq x\leq m_2 \\      U[-1.2,-1] & m_1\ge x\ge m_2 \end{cases}, \\
w(x,0) &\sim\begin{cases} U[-0.1,0.1] & x\leq m_1 \\ U[-0.6,-0.4] & x>m_1 \end{cases},
\end{align*}
where $m_1\sim U[0,1.25]$ and $m_2 \sim m_1+U[0.25,1.5]$, solving the system numerically using a high order method, and observing at time step $\Delta=0.25$ of total time length of $T=50$. The $\sim$ notation along with $U[\cdot,\cdot]$ indicates that this value is drawn from a uniform distribution between the two values. A network structure as in \eqref{NN_simple} (also Figure \ref{fig:pde_structure}) with $5$ disassembly channels each with $1$ hidden layer and $100$ neurons and an assembly layer with $1$ hidden layer and $5$ neurons is used, with outer recurrence $R_{out} = 3$ and inner recurrence $R_{in} = 5$ such that the fine time learning occurs on the scale of $\delta = \Delta/R_{in} = 0.05$. See \cite{chen2022deep} for more details on this structure and its mathematical properties. The mean squared loss function is minimized using Adam with a constant learning rate of $10^{-3}$ for $10,000$ epochs.

\begin{figure}[htbp]
	\begin{center}
		\includegraphics[width=\textwidth]{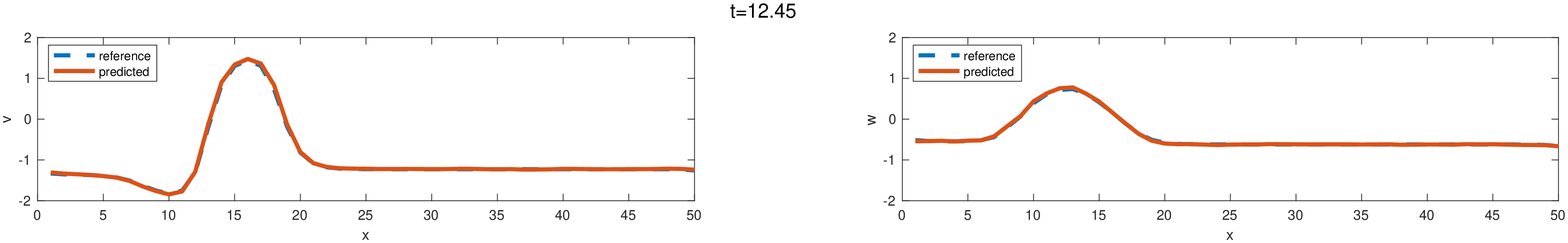}
		\includegraphics[width=\textwidth]{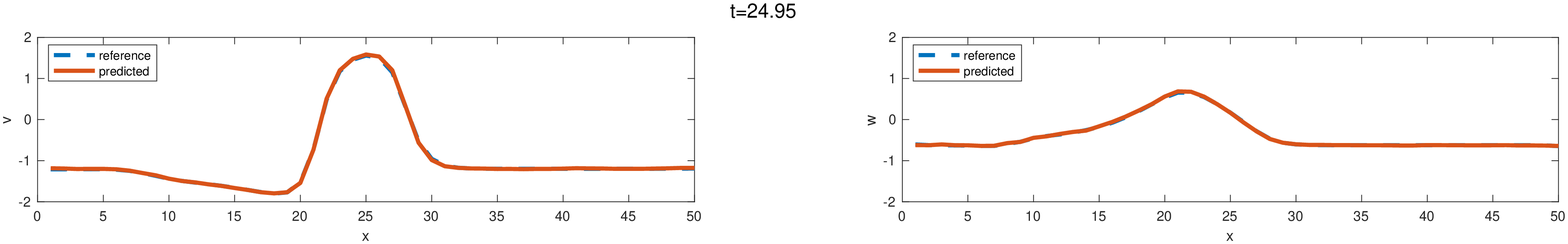}
		\includegraphics[width=\textwidth]{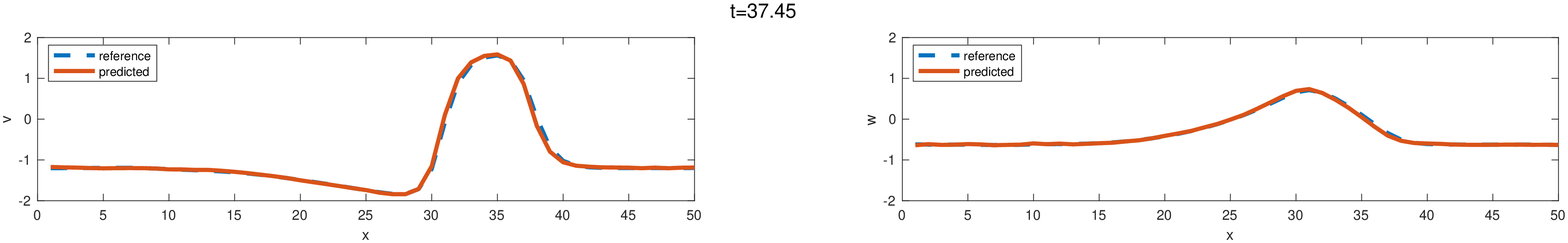}
		\includegraphics[width=\textwidth]{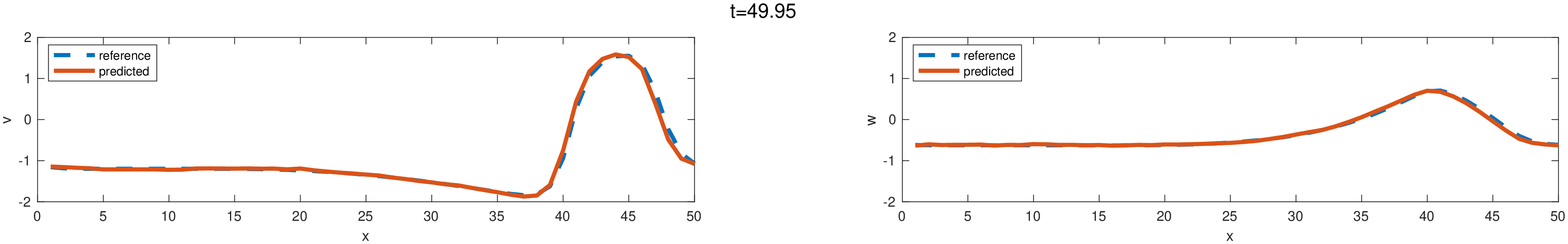}
		\caption{Ex. 5: FitzHugh-Nagumo with diffusion - Example test trajectory.}
		\label{fig:fhn}
	\end{center}
\end{figure}

Prediction is carried out for $T=50$ via $\delta=0.05$ ($2000$ time steps) using $100$ new initial conditions uniformly sampled from the same domain as well as a demonstrative initial condition defined for $x\in[0,5]$ by
\begin{align*}
v(x,0) &= \begin{cases} 1 & 0.75\le x\le1 \\ -1.1 & \text{else} \end{cases} \\
w(x,0) &= \begin{cases} 0 & x\le 0.75 \\ -0.5 & \text{else}. \end{cases}
\end{align*}
The results are shown in Figure \ref{fig:fhn}. We see a strong match between the reference and prediction even after the wave has evolved for a long period of time. We show the prediction at times ($t=12.45$, $24.95$, $37.45$, $49.95$ seconds) exclusively on the $\delta$ time scale (as opposed to a time also coinciding with the $\Delta$ time scale) to demonstrate accuracy on this finer scale.

\subsection{Example 6: Two-dimensional PDE}

As a final ambitious example, we consider the two-dimensional advection-diffusion equation,
\[
	\frac{\partial u}{\partial t} + \nabla\cdot (\boldsymbol{\alpha}u) = \nabla \cdot(\kappa\nabla u)
\]
discretized on a $16\times16$ uniform grid grid over domain $(x,y)\in(-1,1)^2$ with zero Dirichlet boundary conditions. The transport velocity field is set as $\alpha(x,y) = (y,-x)^T$, and the viscosity is $\kappa=5\times10^{-3}$.
Training data are collected from the equation by sampling $100,000$ initial conditions
\[
u_0(x,y) = \sum_{k=1}^{N_c}\sum_{l=1}^{N_c} c_{k,l} \sin\left(\frac{k\pi}{2}(x+1)\right)\sin\left(\frac{l\pi}{2}(y+1)\right),
\]
where $N_c=7$ and the coefficient $c_{k,l}\sim \frac{1}{k+l}U[-1,1]$, and observing a numerical solution at time step $\Delta=0.01$ of total time length of $T=0.1$. A neural network with structure as in \eqref{NN_simple} is used with 3 disassembly channels each with 1 hidden layer consisting of 256 neurons, and an assembly with 1 hidden layer and 3 neurons. The outer recurrence is $R_{out} = 1$ and inner recurrence is $R_{in} = 5$ such that the fine time learning occurs on the scale of $\delta = \Delta/R_{in} = 0.002$. The mean squared loss function is minimized using Adam with a constant learning rate of $10^{-3}$ for $10,000$ epochs.

\begin{figure}[htbp]
	\begin{center}
		\includegraphics[width=.9\textwidth]{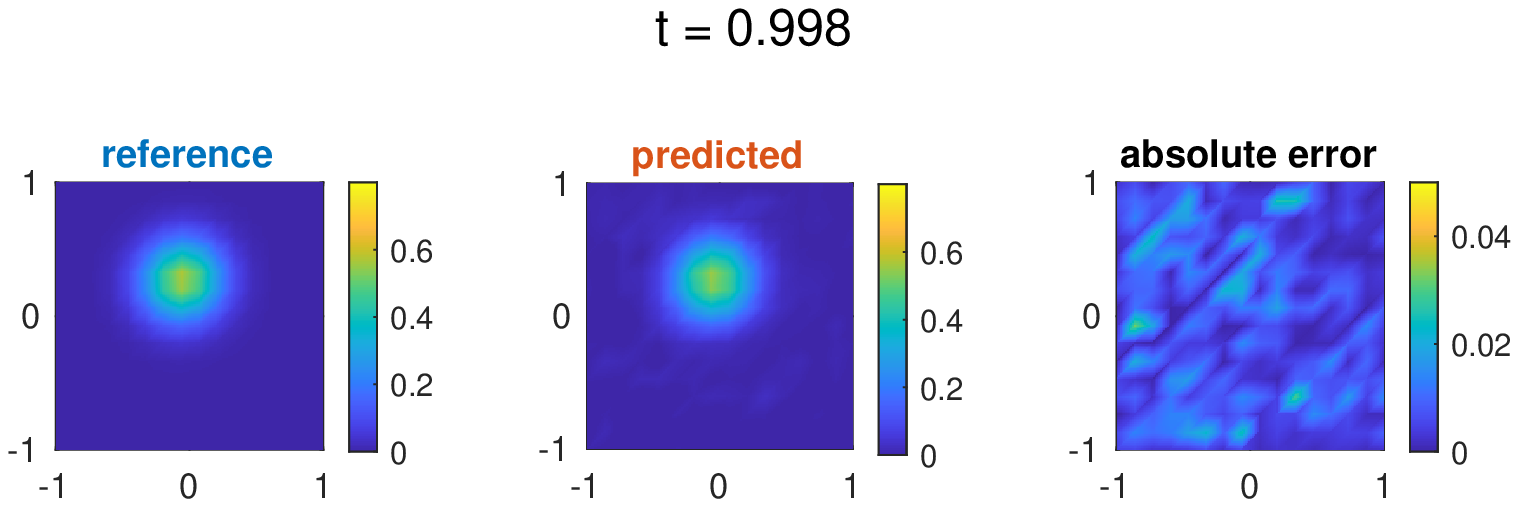}
		\includegraphics[width=.9\textwidth]{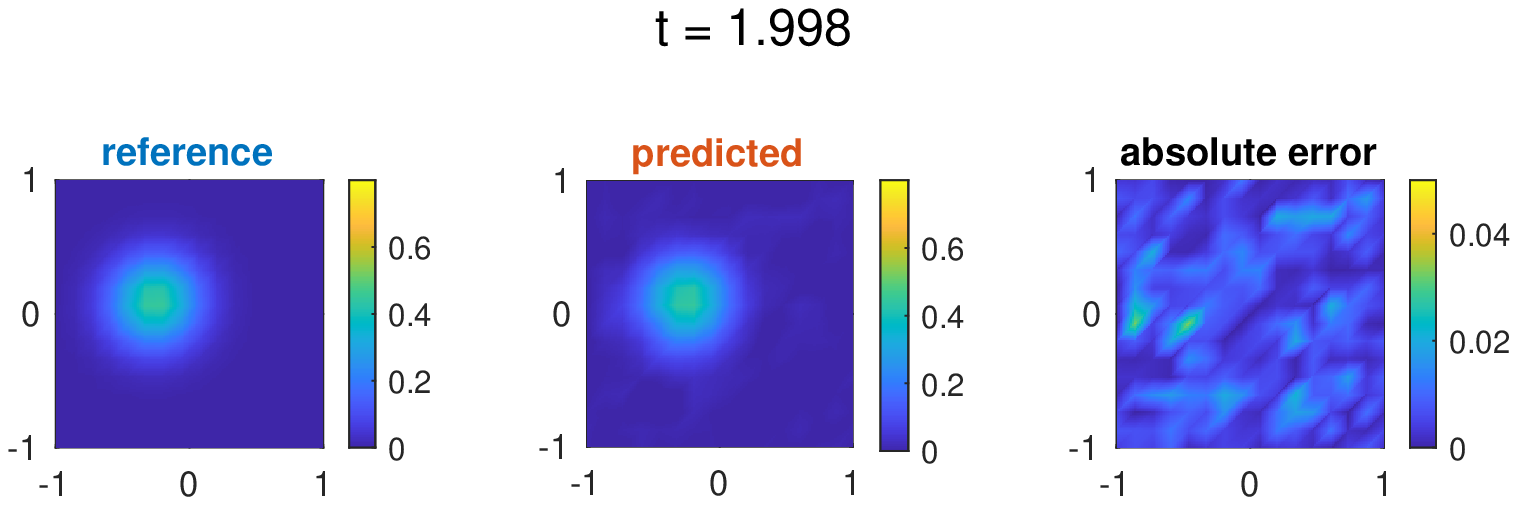}
		\includegraphics[width=.9\textwidth]{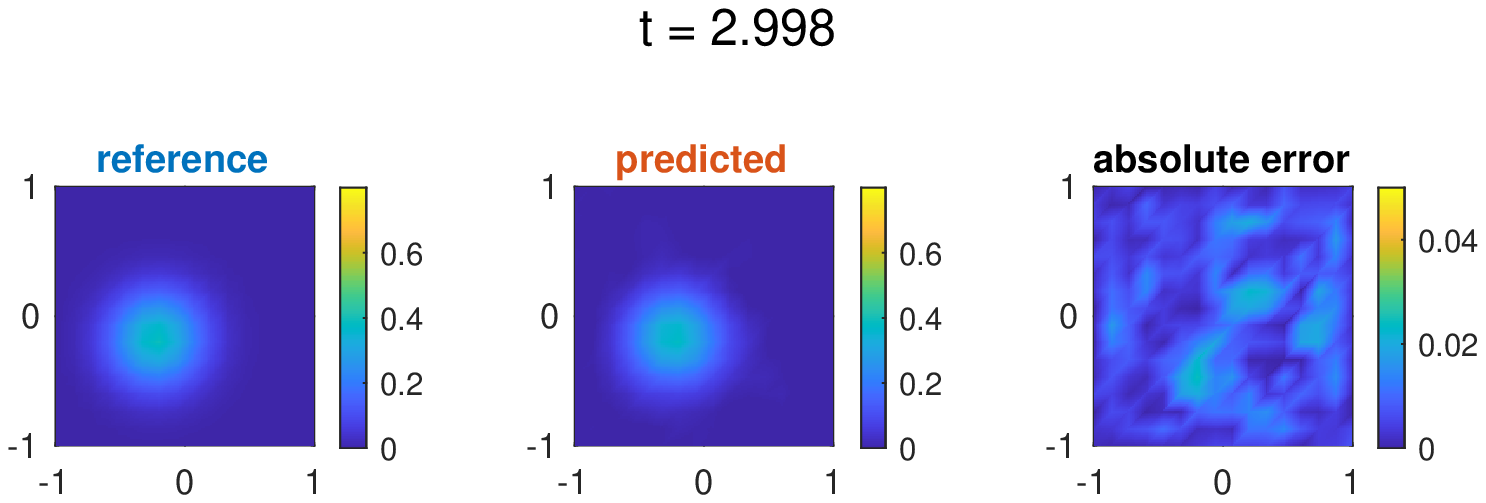}
		\includegraphics[width=.9\textwidth]{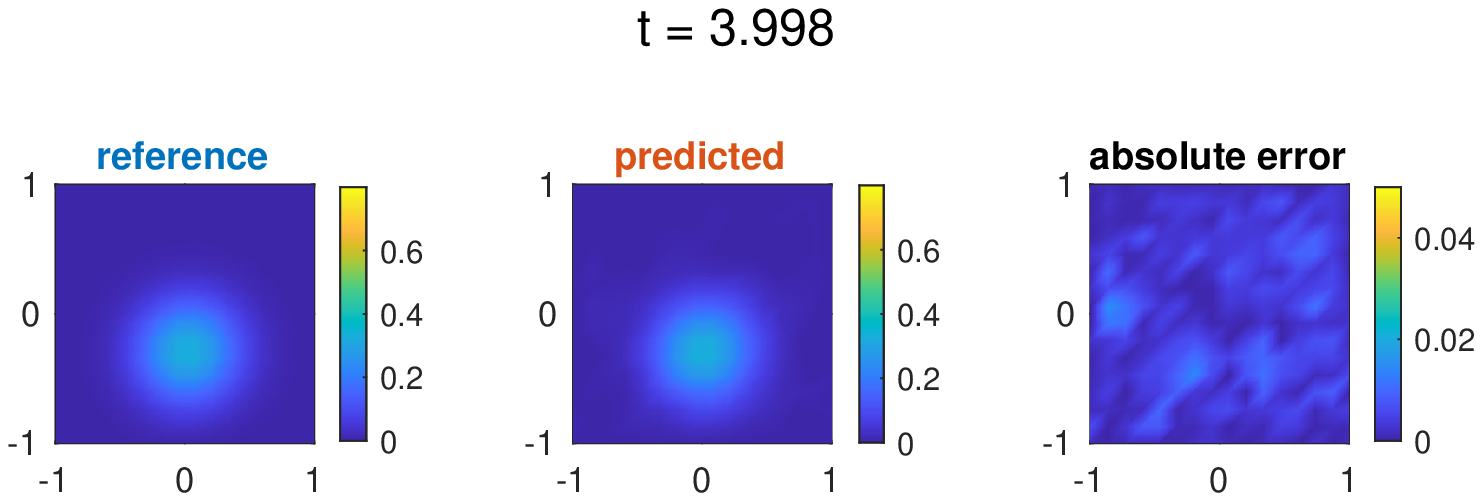}
		\caption{Ex. 6: Two-dimensional PDE.}
		\label{fig:pde}
	\end{center}
\end{figure}

Prediction is carried out for $T=4$ via $\delta=0.002$ ($2000$ time steps) using the initial condition
\[
u_0(x,y) = \frac{C}{2\pi\sigma_x\sigma_y}\exp\left[ -\frac12 \frac{(x-\mu_x)^2}{\sigma_x^2} - \frac12 \frac{(y-\mu_y)^2}{\sigma_y^2}\right],
\]
where $C=0.2$, $\mu_x=\mu_y=0.2$, and $\sigma_x=\sigma_y=0.18$. The qualitative behavior of this example test trajectory is a Gaussian cone revolving around the origin counter-clockwise while diffusing. The results for prediction up to $T=4$ are shown in Fig. \ref{fig:pde}, where we see a strong match for a significantly longer time period than the training data sequences. As above, the times displayed are only on the $\delta$ scale. Interpolatory shading is used to better view results on the fairly coarse $16\times16$ grid.

\section{Conclusion} \label{sec:conclusions}

We have presented a method for fine time-scale DNN learning of unknown systems from trajectories observed on a coarse time scale using the concept of inner recurrence. This new addition builds upon the existing framework for flow map approximation by DNNs and opens it up to real-world problems where data is often sparsely measured in time. A set of six examples are presented which confirms the properties predicted by the foundation and shows that the proposed approach is able to robustly model a variety of standard ODE and PDE problems from data.

\bibliographystyle{siamplain}
\bibliography{neural,LearningEqs,ensemble,lorenz}

\end{document}